\documentclass[letterpaper, 10 pt, conference]{ieeeconf}

\IEEEoverridecommandlockouts                              

\usepackage[pdftex]{graphicx}
\usepackage{amsmath}
\usepackage{booktabs}
\usepackage[nolist, nohyperlinks]{acronym}
\usepackage{amssymb}
\usepackage{bm}
\usepackage{color}
\usepackage{multicol}
\usepackage{multirow}
\usepackage{diagbox}
\usepackage{mathtools}

\begin{document}
\title{\LARGE \bf
An Illumination-Robust Feature Extractor Augmented by Relightable 3D Reconstruction}

\author{Shunyi Zhao,
        Zehuan Yu, \IEEEmembership{Graduate Student Member, IEEE},
        Zuxin Fan,
        Zhihao Zhou, \IEEEmembership{Member, IEEE}\\
        Lecheng Ruan,
        and Qining Wang,~\IEEEmembership{Senior Member, IEEE}
\thanks{Corresponding authors: Lecheng Ruan and Qining Wang.}
\thanks{Shunyi Zhao, Zuxin Fan, and Lecheng Ruan are with the College of Engineering, Peking University, Beijing, 100871, China.}
\thanks{Zehuan Yu is with the Department of Electronic and Computer Engineering, Hong Kong University of Science and Technology, Hong Kong.}
\thanks{Zhihao Zhou is with the Institute for Artificial Intelligence, Peking University, Beijing, 100871, China, and also with the Beijing Engineering Research Center of Intelligent Rehabilitation Engineering, Beijing, 100871, China.}
\thanks{Qining Wang is with the Department of Advanced Manufacturing and Robotics, College of Engineering, Peking University, Beijing 100871, China, with the University of Health and Rehabilitation Sciences, Qingdao, China, with the Institute for Artificial Intelligence, Peking University, Beijing 100871, China.}}

\begin{acronym}
    \acrodef{mse}[MSE]{Mean Square Error}
    \acrodef{bce}[BCE]{Binary Cross Entropy}
    \acrodef{dl}[DL]{Deep Learning}
    \acrodef{nn}[NN]{Neural Network}
    \acrodef{bev}[BEV]{Bird's Eye View}
    \acrodef{lidar}[LiDAR]{Light Detection and Ranging}
    \acrodef{vr}[VR]{Virtual Reality}
    \acrodef{slam}[SLAM]{Simultaneous Localization and Mapping}
    \acrodef{nerf}[NeRF]{Neural Radiance Fields}
    \acrodef{sfm}[SfM]{Structure from Motion}
    \acrodef{ml}[ML]{Machine Learning}
    \acrodef{pca}[PCA]{Principal Components Analysis}
    \acrodef{ae}[AE]{Auto Encoder}
    \acrodef{svd}[SVD]{Singular Value Decomposition}
    \acrodef{cs}[CS]{Cosine Similarity}
    \acrodef{ms}[M. Score]{Matching Score}
    \acrodef{nnmap}[NN mAP]{Nearest Neighbor mean Average Precision}
    \acrodef{nms}[NMS]{Non-Maximum Suppression}
    \acrodef{brdf}[BRDF]{Bidirectional Reflectance Distribution Function}
    \acrodef{nmmap}[NN mAP]{Nearest Neighbour mean Average Precision}
    \acrodef{sfm}[SfM]{Structure from Motion}
    \acrodef{brdf}[BRDF]{Bidirectional Reflectance Distribution Function}
    \acrodef{sp}[SP]{Same-Position}
    \acrodef{dp}[DP]{Different-Position}
    \acrodef{pc}[PC]{Personal Computer}
    \acrodef{pbr}[PBR]{Physical-Based Rendering}
    \acrodef{mvs}[MVS]{Multi-View Stereo}
    \acrodef{map}[mAP]{mean Average Precision}
    \acrodef{sota}[SOTA]{State-Of-The-Art}
\end{acronym}

\def\eg{\emph{e.g.}}
\def\Eg{\emph{E.g.}}
\def\ie{\emph{i.e.}}
\def\Ie{\emph{I.e.}}
\def\etc{\emph{etc.}}

%

\maketitle

\begin{abstract}
Visual features, whose description often relies on the local intensity and gradient direction, have found wide applications in robot navigation and localization in recent years. However, the extraction of visual features is usually disturbed by the variation of illumination conditions, making it challenging for real-world applications. Previous works have addressed this issue by establishing datasets with variations in illumination conditions, but can be costly and time-consuming. This paper proposes a design procedure for an illumination-robust feature extractor, where the recently developed relightable 3D reconstruction techniques are adopted for rapid and direct data generation with varying illumination conditions. A self-supervised framework is proposed for extracting features with advantages in repeatability for key points and similarity for descriptors across good and bad illumination conditions. Experiments are conducted to demonstrate the effectiveness of the proposed method for robust feature extraction. Ablation studies also indicate the effectiveness of the self-supervised framework design.
\end{abstract}


\IEEEpeerreviewmaketitle

\section{Introduction}

Features have garnered considerable attention as a prominent subject of research \cite{lowe2004distinctive, rusu2009fpfh, rublee2011orb, liu2023bevfusion} and have found extensive applications in domains, \eg, autonomous driving \cite{wang2023centernet}, indoor navigation \cite{huang2023visual}, and robot localization \cite{sarlin2023orienternet} in the recent years. Among various categories of features, visual features exhibit distinct advantages in inherent richness in real-world contexts \cite{kurz2014absolute}, ease of acquisition \cite{lee2013embedded}, and incorporation of semantic information \cite{liu2023bevfusion}, thus have attracted significant scholarly scrutiny and have been the subject of extensive investigation.

\begin{figure}
    \centering
    \includegraphics[width=0.41\textwidth]{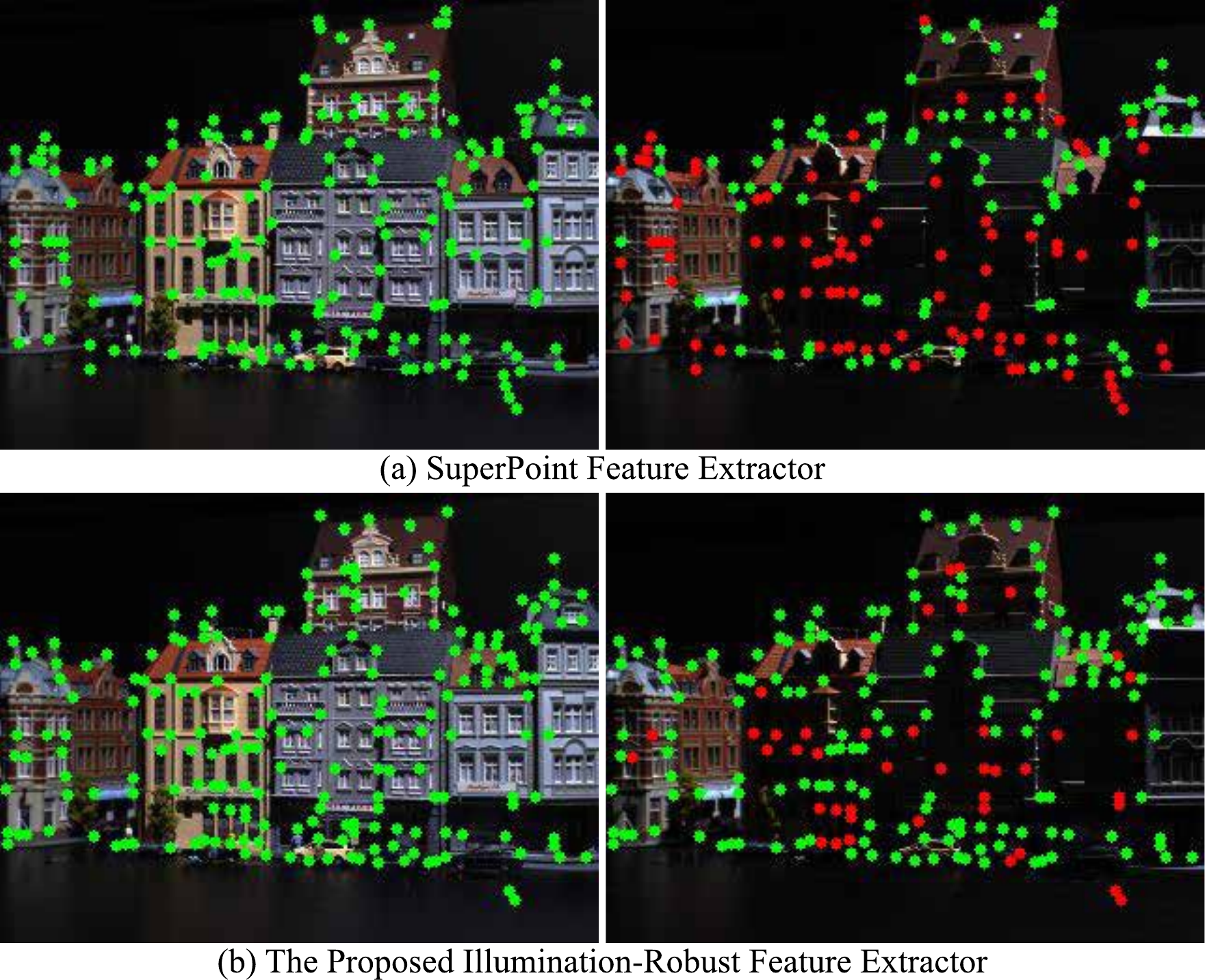}
    \caption{Feature extraction performance tends to degrade with worse illumination conditions (right column compared to left, where red point indicates the missing features). The illumination-robust feature extractor can alleviate such degradation. As an example, the proposed method in (b) alleviates the drain of feature points (repeatability improves from 50.85\% to 72.02\%) and mismatching of descriptors (\ac{map} of descriptor improves from 0.719 to 0.853) compared with the SuperPoint method in (a).}
    \label{fig:problem}
\end{figure}

Description of visual features usually relies on the local intensity \cite{rublee2011orb} or gradient direction \cite{lowe2004distinctive} of pixels as their foundation. Through the fusion of intensity or gradient direction across multiple dimensions, visual feature extractors can localize key points associated with visual features and generate feature descriptors for them. However, the variations in illumination, including photo source orientation, intensity, and color, impose additional challenges for feature extraction \cite{oren1994generalization}, as these variations can disturb the shadows, as well as the color and intensity of reflection lights, and further influence the local intensity and gradient directions of pixels. As a consequence, (1) specific features can vanish, and new plausible features may occur, causing a reduction of \textit{repeatability} \cite{yi2016lift} with varying illumination conditions; (2) even for features that remain active across different illumination conditions, the descriptor can be subject to changes, \ie, the \textit{similarity} of the same feature in different illumination conditions can decrease, which often results in failures for feature matching \cite{yi2016lift, sarlin2020superglue}. Fig. \ref{fig:problem}(a) illustrates the degradation of the feature extraction performance of SuperPoint \cite{detone2018superpoint}, which is a widely-used feature extractor in real-world applications \cite{Xiao2024ARR}, with deteriorated illumination condition.

Great efforts have been made to investigate feature extraction under varying illumination conditions. Conventional feature extractors usually conduct sophisticated designs for particular targeted features with specific domain knowledge to attenuate the effects of illuminations, \eg, face features \cite{kao2010local, nabatchian2011illumination}. Recently, \ac{nn} has been introduced for illumination-robust feature extraction with the intention to improve generality and reduce hand-crafted efforts, either by re-rendering input images to the same illumination condition \cite{anoosheh2019night, mueller2019image} as pre-processing, or directly learning illumination-robust features from large dataset \cite{chen2020if}. However, deep-learning methods inherently require large and representative datasets to demonstrate variations in illuminations, thus could be of heavy workload with artificial lighting equipment \cite{liu2023openillumination}, or may require a long time to collect all desired illumination conditions with natural lighting \cite{balntas2017hpatches, 2020Long}. Using \ac{nn} for data augmentation is one promising direction \cite{venator2021self, musallam2024self}, but to precisely control the illumination condition in directed paths for full coverage of data space, and to also keep the features totally unchanged under different rendering is still challenging due to the inexplicability nature of deep models. 

\begin{figure*}[t]
    \centering
    \includegraphics[width=0.9\textwidth]{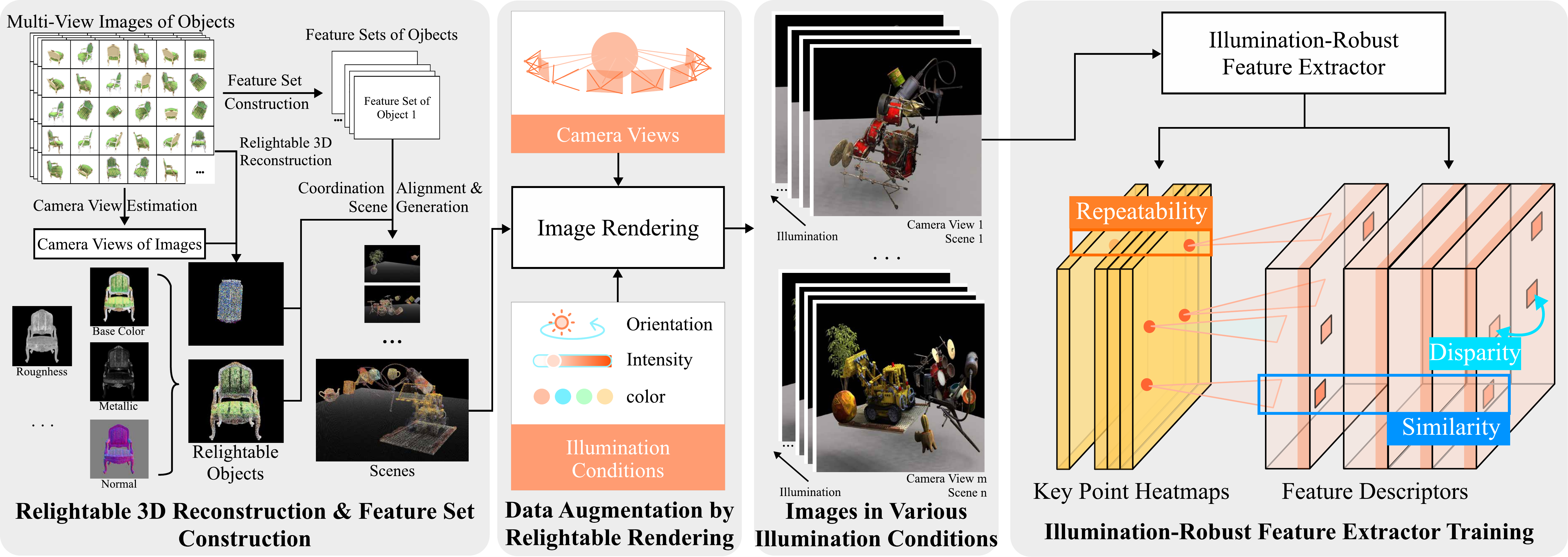}
    \caption{Training pipeline of the proposed illumination-robust feature extractor.}
    \label{fig:pipeline}
\end{figure*}

Recently, relightable 3D reconstruction techniques, grounded in the principles of \ac{nerf} \cite{rudnev2022nerf}, may offer alternative approaches for the aforementioned data augmentation task for illumination robustness. Specifically, these methods can reconstruct 3D scenes with a small set of images, and rapidly render images under different illumination conditions even if the images for reconstruction do not cover the desired illumination conditions. In addition, the illumination conditions can be explicitly controlled, so that the augmentation can be precisely controlled along directed conditions \cite{jin2023tensoir, gao2023relightable}. 

In this paper, we propose a novel illumination-robust feature extractor, based on data augmentation with relightable 3D reconstruction. The augmented data with various illumination conditions and camera views are rendered from the 3D Scenes composed of 3D Gaussian points \cite{kerbl20233d} extended with \ac{brdf} parameters \cite{gao2023relightable}. A self-supervised framework is proposed to address the possible reduction of repeatability and similarity of features between different illumination conditions. The contributions of this paper can be summarized as follows:
\begin{itemize}
    \item [1)] We propose a data generation method that can generate images of multiple objects under diverse illumination conditions. This method enables precise control of illumination conditions based on a relightable 3D reconstruction algorithm.
    \item [2)] We design a self-supervised framework to train an illumination-robust feature extractor. Feature sets are constructed to guide the self-supervised training to promote the repeatability and similarity of features under varying illumination conditions.
    \item [3)] Experiments are conducted to verify the effectiveness of the proposed framework on the improvements of feature repeatability and similarity in varying illumination conditions on real-world datasets. Ablation studies are also conducted to demonstrate the rationality of the self-supervised design. 
\end{itemize}

\section{Method}\label{sec:method}
The design of the proposed illumination-robust feature extractor is shown in Fig. \ref{fig:pipeline}. To achieve precisely controlled illumination augmentation of the dataset, we utilized a relightable 3D reconstruction algorithm \cite{gao2023relightable}, which is based on 3D Gaussian points \cite{kerbl3Dgaussians} extended with \ac{brdf} parameters. In the augmentation process, we estimate camera views of object images, construct feature sets for objects, reconstruct relightable 3D objects, align the coordination between feature sets and relightable 3D objects, generate scenes with the relightable 3D objects and feature sets, render images from the scenes under diverse illuminations, and group images according to camera views and scenes. The image groups are utilized to train the illumination-robust feature extractor with three losses: the repeatability loss guides the locations of key points in bad illumination conditions within the neighborhood of the key points in good conditions; the similarity loss tries to regulate the feature descriptors in different illumination conditions close to each other; and the disparity loss tries to separate different features to avoid potential descriptor collapse.

\subsection{Camera View Estimation}
The camera views corresponding to images are essential for relightable 3D reconstruction. Assuming the camera intrinsic parameters $\bm{K}$ are known or have been estimated by camera calibration methods \cite{zhang1999calib}, we can utilize \ac{sfm} algorithms to estimate the camera views accurately. Colmap \cite{schoenberger2016sfm} is an open-source \ac{sfm} software that is popular for estimation accuracy, for which we select Colmap. In this paper, we (1) organize images according to objects as $\{\bm{\mathcal{I}}_{i}^{obj}\}_{i=1}^{n_{obj}}$; and (2) utilize Colmap to estimate the camera views $\{\bm{\mathcal{T}}_{i}^{obj}\}_{i=1}^{n_{obj}}$ corresponding to images for each object. A camera view $T$ comprises two parts, a rotation matrix $\prescript{T}{}{\bm{q}}$ and a translation vector $\prescript{T}{}{\bm{p}}$.

\subsection{Feature Set Construction}
Feature sets $\{\bm{\mathcal{F}}_{i}\}_{i=1}^{n_{obj}}$, comprising the 3D coordinates for feature key points, are constructed to guide the training of the illumination-robust feature extractor. Taking the feature set construction process of the $i$-th object as an example, we (1) calculate a feature superset, which consists of pixel coordinates of feature key points, with the homography adaptation method proposed by \cite{detone2018superpoint} for all the images of the $i$-th object; and (2) project the pixel coordinates to 3D coordinates referring to $\bm{\mathcal{T}}_{i}$ with
\begin{equation}
    \frac{\prescript{3D}{}{\bm{\zeta}}}{z}
     = \begin{bmatrix}
        \prescript{T}{}{\bm{p}}_{j}^{i} & \prescript{T}{}{\bm{q}}_{j}^{i}
    \end{bmatrix}^{-1} \bm{K}^{-1}\prescript{}{}{\bm{\zeta}},
    \label{eq:2Dto3D}
\end{equation}
where $\bm{\zeta}$ denotes a pixel coordinate in the $j$-th image, $\prescript{3D}{}{\bm{\zeta}}$ denotes the 3D coordinate, $\prescript{T}{}{\bm{p}}_{j}^{i}$ and $\prescript{T}{}{\bm{q}}_{j}^{i}$ are the estimated camera view of the $j$-th image, and $z$ is the depth measurement of $\bm{\zeta}$ which is a intermediate result in the camera view estimation. By processing all objects, the feature sets $\{\bm{\mathcal{F}}_{i}\}_{i=1}^{n_{obj}}$ can be constructed.

\subsection{Relightable 3D Object Reconstruction}
This part aims to reconstruct relightable 3D objects, which can be utilized in image rendering under different illumination conditions. \cite{gao2023relightable} can reconstruct relightable 3D objects and is famous for the rendering accuracy and short reconstruction and rendering time. Given images $\bm{\mathcal{I}}_{i}^{obj}$ and corresponding camera views $\bm{\mathcal{T}}_{i}^{obj}$ of the $i$-th object, the objective of \cite{gao2023relightable} is to reconstruct a 3D object $\bm{O}_{i}$ composed of 3D Gaussian points $\bm{\mathcal{G}}_{i}=\{G_{j}^{i}\}_{j=1}^{n_{\bm{\mathcal{G}}_{i}}}$ extended with \ac{brdf} parameters. The basic components of a 3D Gaussian point can be represented as $G(\bm{p})=e^{-\frac{1}{2}\bm{p}^{T}\bm{\Sigma}\bm{p}}$, where $\bm{p}$ denotes the position of $G$ and $\bm{\Sigma}$ denotes the covariance of $G$ which controls the shape of $G$, and an opacity $o$. The \ac{brdf} parameters, including base color $\bm{b}$, roughness $r$, metallic properties $\bm{m}$, and spatial normal $\bm{n}$, are added to the basic components of $G$ to enable relightable rendering. Given $\bm{\mathcal{G}}_{i}$, a \ac{pbr} method can be utilized to render an RGB image $\prescript{r}{}{I}$ and generate a depth image $\prescript{rd}{}{I}$ with a preset camera view. The reconstruction minimizes three differences, including (1) the difference between the input images and the rendered images from the same camera views; (2) the difference between the estimated normal and normal calculated by \ac{mvs}; and (3) the difference between the generated depth image $\prescript{rd}{}{I}$ and the depth image calculated by \ac{mvs}, by tuning adjusting the parameters of $\bm{\mathcal{G}}_{i}$. During the optimization process, the number of Gaussian points in $\bm{\mathcal{G}}_{i}$ is also adjusted by two conditions, whose details are referred to \cite{kerbl3Dgaussians}. By applying this reconstruction method to all objects, relightable 3D objects $\bm{\mathcal{O}}=\{\bm{O}_{i}\}_{i=1}^{n_{\bm{\mathcal{O}}}}$ can be acquired.  

\subsection{Coordination Alignment \& Scene Generation}
The coordination between the feature sets $\bm{\mathcal{F}}$ and relightable 3D objects $\bm{\mathcal{O}}$ are aligned, and the aligned set-object pairs are utilized to compose to generate various scenes to realize more complicated image rendering to guarantee the variety of the augmented dataset. Given the feature set $\bm{\mathcal{F}}_{i}$ and the relightable 3D object $\bm{O}_{i}$ of the $i$-th object, the coordination alignment is a union operation. The aligned feature-set-object pair can be represented as $\langle \bm{\mathcal{F}}_{i}, \bm{O}_{i} \rangle$. 

Scenes are generated by composing different feature-set-object pairs. Let $\bm{\mathcal{P}} = \{\bm{P}_{i}:=\langle \bm{\mathcal{F}}_{i}, \bm{O}_{i} \rangle\}_{i=1}^{n_{\bm{\mathcal{P}}}}$ denotes the pairs, we (1) generate a rotation matrix and a translation vector randomly for each pair, and (2) compose the objects by
\begin{equation}
    \small
    \bm{S} = \bigcup_{i=1}^{n_{\bm{\mathcal{P}}}}\begin{bmatrix}
        \prescript{3\times 3}{}{\bm{R}_{i}} & \prescript{3\times 1}{}{\bm{t}}_{i} \\
        \prescript{1\times 3}{}{\bm{0}} & 1
    \end{bmatrix}\begin{bmatrix}
        \bm{P}_{i}
    \end{bmatrix}_{\bm{p}},
    \label{eq:scene_gene}
\end{equation}
where $\prescript{3\times 3}{}{\bm{R}}_{i}$ denotes the random rotation matrix and $\prescript{3\times 1}{}{t}_{i}$ denotes the random translation vector. The operation $\begin{bmatrix}
    \cdot
\end{bmatrix}_{\bm{p}}$ denotes that
\begin{equation}
    \small
    \begin{bmatrix}
        \bm{P}_{i}
    \end{bmatrix} = \begin{bmatrix}
        \begin{matrix}
            \prescript{3\times 1}{}{\bm{p}}^{1} \\
            1
        \end{matrix} & ... & \begin{matrix}
            \prescript{3\times 1}{}{\bm{p}}^{n_{\bm{P}_{i}}} \\
            1
        \end{matrix}
    \end{bmatrix},
\end{equation}
where $n_{\bm{P}_{i}}$ denotes the sum of number of 3D Gaussian points and feature points in the pair $\bm{P}_{i}$. By randomly selecting different objects, various scenes $\bm{\mathcal{S}}=\{\bm{S}_{i}\}_{i=1}^{n_{\bm{\mathcal{S}}}}$, different in object numbers, positions, and rotations, can be generated.

\subsection{Data Augmentation by Relightable Rendering}
Images under various illumination conditions are rendered from $\bm{\mathcal{S}}$ to make up an augmented dataset to train the illumination-robust feature extractor. The dataset augmentation process comprises three steps, including (1) setting camera views $\prescript{r}{}{\bm{\mathcal{T}}}$; (2) defining illumination conditions $\bm{\mathcal{E}}$ obeying the rules in \cite{physg2021}; and (3) for each composition of camera view, illumination condition, and scene, rendering a image with the \ac{pbr} and projecting the feature sets to the rendered image. After the data augmentation process, we group the images and projected feature sets according to the same camera view and scene. The groups can be represented as $\{\langle \prescript{rg}{}{ \bm{\mathcal{I}}}_{i}, \prescript{rg}{}{\bm{\mathcal{F}}}_{i}\rangle \}_{i=1}^{n_{\bm{\mathcal{S}}}\times n_{\prescript{r}{}{\bm{\mathcal{T}}}}}$.

\subsection{Illumination-Robust Feature Extractor Training}
The illumination-robust feature extractor, which addresses feature repeatability and similarity, is trained with the augmented dataset $\{\prescript{rg}{}{\bm{\mathcal{I}}}_{i}\}_{i=1}^{n_{\bm{\mathcal{S}}}\times n_{\prescript{r}{}{\bm{\mathcal{T}}}}}$. We outline the structure of the feature extractor and the self-supervised training framework in this part. The structure of the feature extractor is similar to \cite{detone2018superpoint}, comprising an encoder $f^{e}(\cdot)$, a key point decoder $f^{k}(\cdot)$, and a descriptor decoder $f^{d}(\cdot)$. The $f^{e}(\cdot)$ takes a gray-scale image whose shape is $H\times W\times 1$ as input, outputs a embedding $\prescript{m}{}{\bm{D}}$. The $f^{k}(\cdot)$ and $f^{d}(\cdot)$ process $\prescript{m}{}{\bm{D}}$ separately, output a key point heat map $\prescript{k}{}{\bm{D}}$ whose shape is $\frac{H}{8}\times \frac{W}{8}\times 65$ which has a channel representing dustbin and a feature descriptor map $\prescript{d}{}{\bm{D}}$ whose shape is $H\times W\times 256$. 

The training process of the illumination-robust feature extractor is formulated as a self-supervised framework to address the feature repeatability and similarity. With the $i$-th group of rendered images and projected features $\langle \prescript{rg}{}{\bm{\mathcal{I}}}_{i}, \prescript{rg}{}{\bm{\mathcal{F}}}_{i} \rangle$, the feature extractor can generate key point heatmaps $\{\prescript{k}{}{\bm{D}}_{j}\}_{j=1}^{n_{\bm{\mathcal{E}}}}$ and descriptor maps $\{\prescript{d}{}{\bm{D}}_{j}\}_{j=1}^{n_{\bm{\mathcal{E}}}}$. With $\{\prescript{k}{}{\bm{D}}_{j}\}_{j=1}^{n_{\bm{\mathcal{E}}}}$, we set a repeatability loss as
\begin{equation}
    \small
    \mathcal{L}^{r}(\{\prescript{k}{}{\bm{D}}_{j}\}_{j=1}^{n_{\bm{\mathcal{E}}}}, 
 \bm{\mathcal{Y}}_{i}) = \frac{1}{n_{\bm{\mathcal{E}}}} \sum_{j=1}^{n_{\bm{\mathcal{E}}}} l^{r}(\prescript{k}{}{\bm{D}}_{j}, \bm{\mathcal{Y}}_{i}),
\end{equation}
where $\bm{\mathcal{Y}}_{i}$ is a key point heat map, whose calculation method is reshaping the $\prescript{rg}{}{\bm{\mathcal{F}}}_{i}$ to $\frac{H}{8}\times \frac{W}{8}\times 64$ and adding a dustbin channel. The $l^{r}(\cdot)$ is a cross-entropy loss which can be denoted as
\begin{equation}
    \small
    l^{r}(\prescript{k}{}{\bm{D}}_{j}, \bm{\mathcal{Y}}_{i}) = \sum_{m=1,n=1}^{H,W} -\log \frac{\exp{\prescript{k}{}{\bm{D}}_{j}^{m, n, y}}}{\sum_{h=1}^{65}\exp{\prescript{k}{}{\bm{D}}_{j}^{m, n, h}}},
    \label{equ:rep_cross}
\end{equation}
where $y$ is the channel number where the key point is active in $\bm{\mathcal{Y}}_{i}$ at the position $(m, n)$. With $\{\prescript{d}{}{\bm{D}}_{j}\}_{j=1}^{n_{\bm{\mathcal{E}}}}$, we set a similarity loss as
\begin{equation}
    \small
    \mathcal{L}^{i}(\{\prescript{d}{}{\bm{D}}_{j}\}_{j=1}^{n_{\bm{\mathcal{E}}}}) = \frac{\sum_{m=1}^{n_{\bm{\mathcal{E}}}}\sum_{n=m+1}^{n_{\bm{\mathcal{E}}}}l^{fu}( \prescript{d}{}{\bm{D}}_{m}, \prescript{d}{}{\bm{D}}_{n})}{(n_{\bm{\mathcal{E}}})!}.
    \label{equ:sim_loss}
\end{equation}
In equation \eqref{equ:sim_loss}, $l^{fu}(\cdot)$ is a fusion error function defined as
\begin{equation}
    l^{fu}(\prescript{d}{}{\bm{D}}_{1}, \prescript{d}{}{\bm{D}}_{2}) = l^{l2}(\prescript{d}{}{\bm{D}}_{1}, \prescript{d}{}{\bm{D}}_{2}) + 1 - l^{cs}(\prescript{d}{}{\bm{D}}_{1}, \prescript{d}{}{\bm{D}}_2),
\end{equation}
where $l^{l2}(\cdot)$ denotes a \ac{mse} and $l^{cs}(\cdot)$ denotes a \ac{cs}. The $l^{l2}(\cdot)$ is defined as
\begin{equation}
    \small
    l^{l2}(\bm{D}_{1}, \bm{D}_{2}) = \frac{1}{HW}\sum_{m=1}^{H}\sum_{n=1}^{W}\frac{1}{C}\sum_{h=1}^{C}(\bm{D}_{1}^{m, n, h}, \bm{D}_{2}^{m, n, h})^{2}.
    \label{equ:mse}
\end{equation}
The $l^{cs}(\cdot)$ is defined as
\begin{equation}
    \small
    l^{cs}(\prescript{d}{}{\bm{D}}_{1}, \prescript{d}{}{\bm{D}}_{2}) = \frac{1}{HW}\sum_{m=1}^{H}\sum_{n=1}^{W} \frac{\prescript{d}{}{\bm{D}}_{1}^{m, n}\cdot \prescript{d}{}{\bm{D}}_{2}^{m, n}}{\lVert \prescript{d}{}{\bm{D}}_{1}^{m, n} \rVert \lVert \prescript{d}{}{\bm{D}}_{2}^{m,n} \rVert }.
    \label{equ:cs}
\end{equation}
To prevent the descriptor collapse caused by \eqref{equ:sim_loss}, we formulate a disparity loss as 
\begin{equation}
\small
    \begin{aligned}
          \mathcal{L}^{d}(\{\prescript{d}{}{\bm{D}}_{j}\}_{j=1}^{n_{\bm{\mathcal{E}}}}) 
        = \frac{1}{n_{\bm{\mathcal{E}}}} \sum_{j=1}^{n_{\bm{\mathcal{E}}}}\frac{\sum_{m=1}^{n_{k}}\sum_{n=m+1}^{n_{k}}l^{fu}(\bm{d}_{m}^{j}, \bm{d}_{n}^{j})}{(n_{k}-1)!},
    \end{aligned}
    \label{equ:loss_disp}
\end{equation}
where $\bm{d}_{m}^{j}$ and $\bm{d}_{n}^{j}$ denote two descriptors in $\prescript{d}{}{\bm{D}}_{j}$ where the key points are active in $\bm{\mathcal{Y}}$. The optimization object of the feature extractor is minimizing $\mathcal{L}^{r}$ and $\mathcal{L}^{i}$ and maximizing $\mathcal{L}^{d}$. Therefore, the loss function can be summarized as
\begin{equation}
    \begin{aligned}
        \mathcal{L} = \lambda_{1}\mathcal{L}^{r}+ \lambda_{2}\mathcal{L}^{i} +\lambda_{3} \frac{1}{\mathcal{L}^{d}}, 
    \end{aligned}
    \label{equ:loss_sum}
\end{equation}
where $\lambda$ is the coefficient to balance the three losses.

\section{Experiment Setup}\label{sec:exp_setup}
\begin{figure*}
    \centering
    \includegraphics[width=0.945\textwidth]{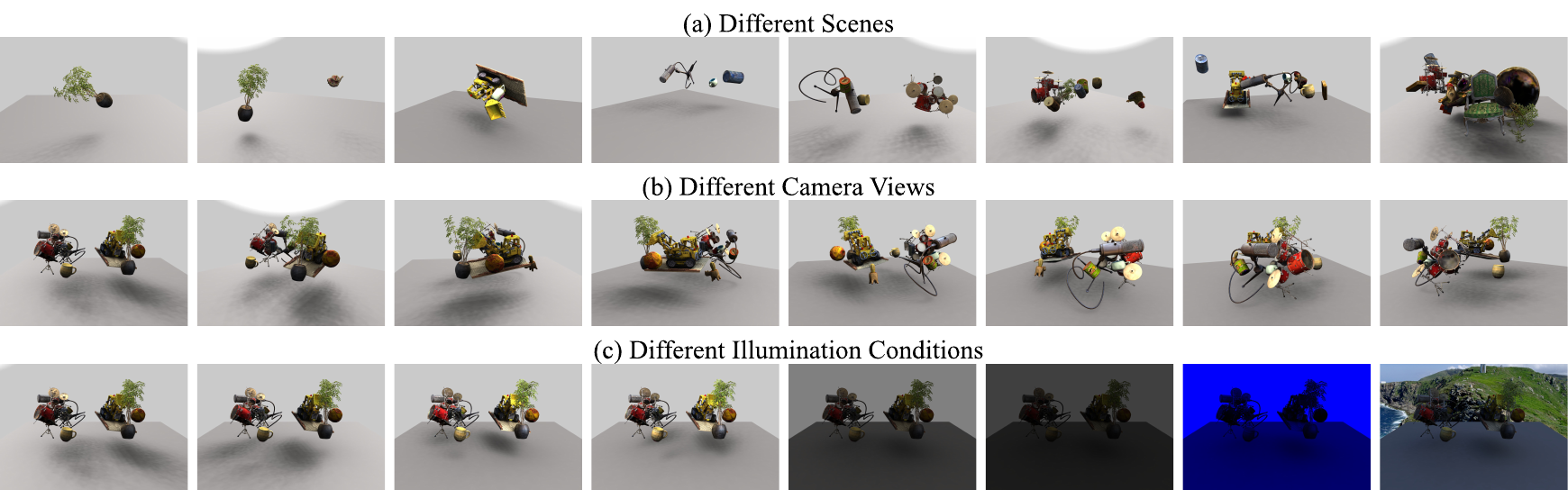}
    \caption{The proposed data augmentation method can generate figures with (a) different scene configurations, (b) different camera views, and (c) different illumination conditions, for the training of the illumination-robust feature extractor.}
    \label{fig:r_images}
\end{figure*}

This section provides information about experimental settings, including the hardware and software environments, the implementation details of the illumination-robust feature extractor, and the experimental protocols.

\subsection{Hardware \& Software Environment}
The experiments are conducted on a \ac{pc} equipped with an Intel Core i9-13900KF, 64 GB memory, and a Nvidia Geforce RTX 4090. PyTorch 2.2.1 and CUDA Toolkit 12.2 are utilized to implement and execute \ac{nn}-based algorithms. Additionally, classical feature extractors, such as SIFT \cite{lowe2004distinctive}, are implemented using the OpenCV framework.

\subsection{Implementation Details of Data Augmentation \& Training}
The data augmentation incorporates four open-source datasets, including \ac{nerf} Synthesis \cite{mildenhall2020nerf}, DTU MVS \cite{jensen2014large}, OpenIllumination \cite{liu2023openillumination}, and Standford-ORB \cite{kuang2024stanford}. These datasets encompass various objects, with approximately 100 images per object for 3D object reconstruction. During the dataset augmentation process introduced in Section \ref{sec:method}, we preset 300 camera views and 13 illumination conditions to render images. COCO \cite{lin2014microsoft}, whose key points are generated by the homography adaption proposed by \cite{detone2018superpoint}, is also utilized to train the feature detector decoder $f^{d}(\cdot)$ to promote the generalizability of the key point detector.

\subsection{Experiment Protocols}

\subsubsection{Feature Extractor Runtime}
We test the average runtime of the proposed illumination-robust feature extractor with the 100 frames of $320\times 240$ images both on CPU and GPU.

\subsubsection{Key Point Repeatability}
The key point repeatability of the proposed illumination-robust feature extractor under diverse illumination conditions is considered a crucial characteristic. To evaluate this, we employ images of three objects from the open-source dataset OpenIllumination \cite{liu2023openillumination} and the varying-illumination images in HPatches \cite{balntas2017hpatches}. These two datasets contain multiple sets of images. Each set is captured from the same camera views under diverse illumination conditions. For every camera view, we first select the brightest image and detect feature key points in this image as reference key points, then calculate the proportion of repeatedly detected key points in other images and the reference key points as repeatability. SIFT \cite{lowe2004distinctive} and SuperPoint \cite{detone2018superpoint}, which is a typacial feature extractor in general, are utilized to compare. The SuperPoint feature extractor we use to compare is trained from scratch with the same training iterations of the proposed illumination-robust feature extractor.

\subsubsection{Feature Descriptor Similarity}
The similarity of \ac{sp} feature descriptors under diverse illumination conditions determines the feature matching accuracy. To evaluate the similarity, we calculate the \ac{mse} and \ac{cs} for descriptors of the repeatedly detected key points in the repeatability test. SIFT \cite{lowe2004distinctive} and SuperPoint \cite{detone2018superpoint} are utilized to compare. 

\subsubsection{Ablation Study}
Except for the repeatability loss which has been validated through the feature detection repeatability test, it is crucial to assess the effectiveness of the remaining two losses within the proposed framework.

\emph{Similarity Loss:}
The similarity loss serves the purpose of ensuring the similarity of the \ac{sp} feature descriptors, under various illumination conditions. We retrain the illumination-robust feature extractor solely using the repeatability and the disparity losses and record the similarity metrics, including \ac{mse} and \ac{cs}, of the \ac{sp} descriptors during the training process. We compare the performance of this feature extractor with another feature extractor trained with three losses.

\emph{Disparity Loss:}
The disparity loss is employed to ensure the disparity of \ac{dp} feature descriptors. To examine the impact of this loss, we retrain the illumination-robust feature extractor using the repeatability and the similarity losses, and test the disparity, including \ac{mse} and \ac{cs}, of \ac{dp} descriptors. We compare the performance of this feature extractor with another feature extractor trained with three losses.

\section{Experiment Result}\label{sec:exp_res}

\subsection{Feature Extractor Runtime}
The average runtimes of the proposed illumination-robust feature extractor are 20.73 ms on GPU and 44.61 ms on CPU, which indicates that the illumination-robust feature extractor can be utilized in real-time applications with a GPU.

\subsection{Feature Detection Repeatability}
The evaluation results of feature detection repeatability are presented in TABLE \ref{tab:eva_res}. From this table, the proposed illumination-robust feature extractor attains a feature repeatability rate of 45.09\% when distance threshold $\epsilon = 1$ with an average location error of 0.31 under diverse illumination conditions, which indicates the proposed method can improve the feature detection repeatability in diverse illuminations.

\begin{table}[!ht]
    \centering
    \caption{Feature Detection Repeatability and Similarity}
    \resizebox{0.44\textwidth}{!}{
    \begin{tabular}{c|cccc}
        \toprule
        Feature Extractor & Repeatability & Location Error & \ac{mse} & \ac{cs} \\
        \midrule
        SIFT & 29.05\% & 0.46 & 1.10 & 0.91 \\
        SuperPoint & 39.61\% & 0.49 & 1.45 $\times$ 10$^{\mathrm{-3}}$ & 0.89 \\
        Ours & 45.09\% & 0.31 & 0.61 $\times$ 10$^{\mathrm{-3}}$ & 0.98 \\
        \bottomrule
    \end{tabular}
    }
    \label{tab:eva_res}
\end{table}

\subsection{Feature Descriptor Similarity}
The evaluation results of the feature descriptor similarity are shown in TABLE \ref{tab:eva_res}. From this table, the illumination-robust feature extractor attains a \ac{mse} value of $0.61 \times 10^{-3}$ and a \ac{cs} value of 0.98, which indicates the improvement of the proposed method in the similarity of \ac{sp} descriptors.

\begin{figure}[t!]
    \centering
    \includegraphics[width=0.41\textwidth]{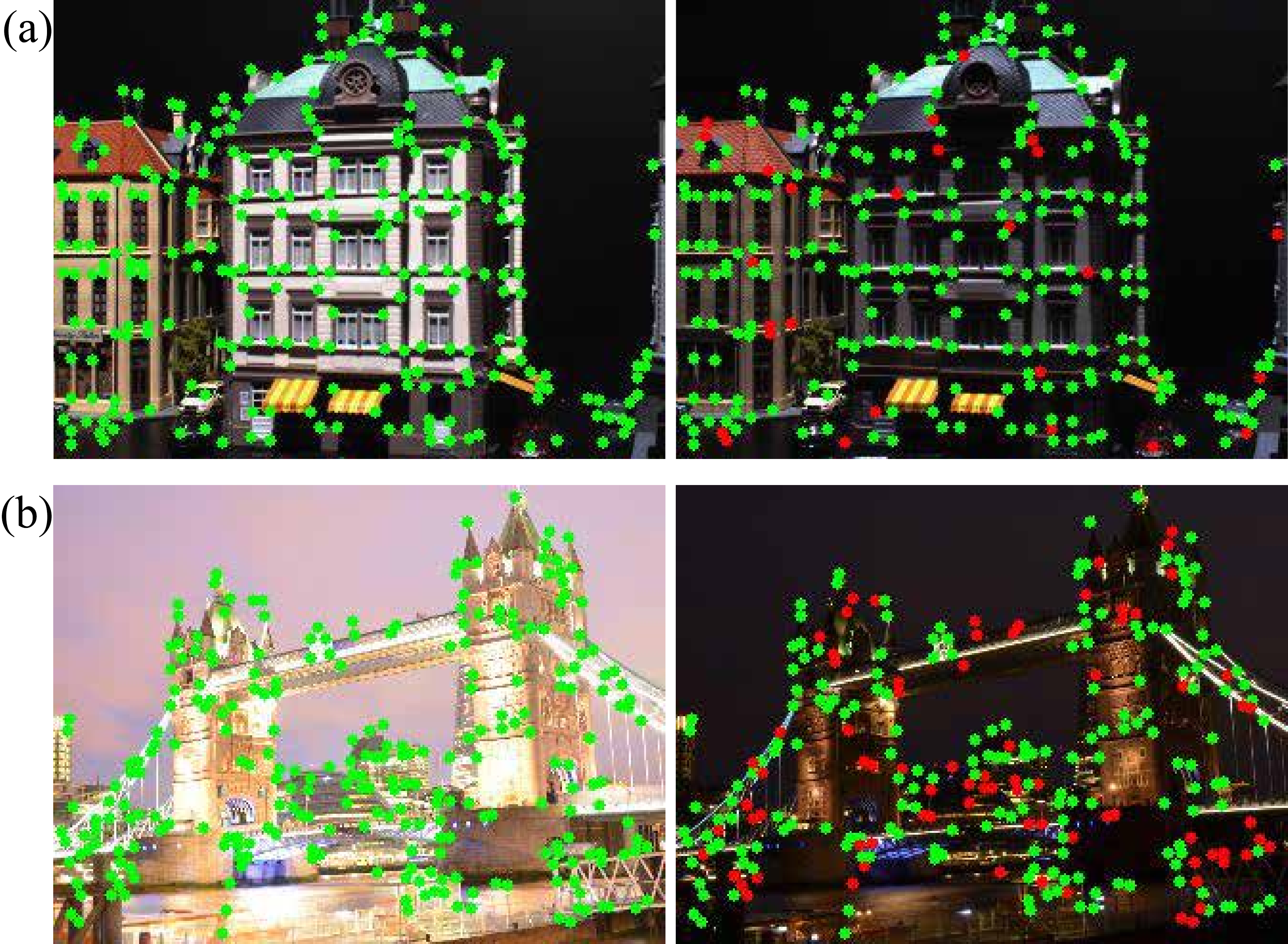}
    \caption{The feature extraction results of the proposed illumination-robust feature extractor across different illumination conditions.}
    \label{fig:matching_image}
\end{figure}

\subsection{Ablation Study}
\subsubsection{Similarity Loss}
The evaluation outcomes of the similarity loss are illustrated in TABLE \ref{tab:ablation}. In the absence of similarity loss, the average \ac{mse} increases from $0.06\times 10^{-2}$ to $0.84\times 10^{-2}$, and the average \ac{cs} decreases from 0.98 to 0.72. These results indicate a decline in the similarity of \ac{sp} features when the similarity loss is not incorporated.

\subsubsection{Disparity Loss}
The evaluation results are presented in TABLE \ref{tab:ablation}. Without the disparity loss, the average \ac{mse} decreases from $2.87\times 10^{-2}$ to $0.48\times 10^{-2}$, and the average \ac{cs} increases from 0.05 to 0.92. These evaluation results suggest the necessity of the disparity loss. 

\begin{table}[!h]
    \centering
    \caption{The Effectiveness of the Similarity and Disparity Losses}
    \resizebox{0.32\textwidth}{!}{
    \begin{tabular}{c|cccc}
        \toprule
        \multirow{3}{*}{Model} & \multicolumn{4}{c}{Metric} \\
        & \multicolumn{2}{c}{\ac{mse}($10^{-2}$)} & \multicolumn{2}{c}{\ac{cs}} \\
        & SP & DP & SP & DP \\
        \midrule
        with both $\mathcal{L}^{i}$ and $\mathcal{L}^{d}$ & 0.06 & 2.87 & 0.98 & 0.05 \\
        only with $\mathcal{L}^{i}$ & 0.02 & 0.48 & 0.99 & 0.92 \\
        only with $\mathcal{L}^{d}$ & 0.84 & 2.89 & 0.72 & 0.04 \\
        \bottomrule
    \end{tabular}
    }
    \label{tab:ablation}
\end{table}

\section{Discussion}\label{sec:diss}
Extraction of visual features, comprising key points and descriptors, is the basis of various robotic applications, such as visual place recognition, automatic drive, and robot navigation. However, variations of illumination conditions challenge the performance, \eg, key point repeatability and descriptor similarity, of visual feature extractors. To enhance the performance of the feature extractors, specific feature designing for specific domains \cite{kao2010local, nabatchian2011illumination}, dataset augmentation by image-illumination transformation \cite{anoosheh2019night, mueller2019image}, and cross-illumination self-supervised feature extractor framework \cite{venator2021self, musallam2024self}, are utilized. However, the diversity of illumination conditions still limits the performance of these works. In this paper, we propose an illumination-robust feature extractor, based on a relightable 3D reconstruction to augment the dataset. Two important evaluation metrics, repeatability and similarity, are addressed by three losses, the repeatability loss, the similarity loss, and the disparity loss, which are used to train the illumination-robust feature extractor.

Firstly, we highlight our data augmentation approach. The performance of previous studies \cite{anoosheh2019night, venator2021self} under different illuminations relies on the scalability and variety of the datasets. But the collection of data under various illumination conditions \cite{2020Long, liu2023openillumination, balntas2017hpatches} in the world is still struggled by (1) the requirement of a large number of images; (2) the complexity and difficulty of artificial construction and precise control of illuminations; and (3) the limitation of specific time periods of outdoor scenes. In this paper, we propose a data augmentation method to realize precise control of the illumination conditions based on a relightable 3D reconstruction algorithm. With this algorithm, 3D objects can be reconstructed by approximately 100 2D RGB images within 40 minutes. After reconstruction, 40 images per minute on average can be rendered for the one-object scenes under diverse illumination conditions and camera views. The illumination conditions can be controlled precisely during the rendering, showcasing the usability. Example images are shown in Fig. \ref{fig:r_images}. These images demonstrate that our data augmentation method enables separate control over object number, object type, illumination orientation, intensity, and color with predetermined camera views.

Secondly, we address the feature repeatability \cite{yi2016lift} and similarity under diverse illumination conditions with a self-supervised training framework. The repeatability loss attempts to guide the feature region under bad illumination conditions within the neighborhood of the features with preferred conditions. The similarity loss tries to regulate the feature descriptors under different illumination conditions close to each other. To avoid potential descriptor collapse caused by the similarity loss, the disparity loss is added. Comprehensive experiments are conducted to verify the performance of this illumination-robust feature extractor. The experiment results demonstrate that the repeatability of feature key point detection achieves 45.09\%, and the similarity of \ac{sp} feature descriptors achieves a \ac{mse} of $0.61\times 10^{-3}$ and a \ac{cs} of 0.98. Compared with previous popular feature extractors, improvements can be observed. Ablation studies are also conducted to emphasize the importance of the similarity and disparity losses. Utilizing different loss functions (both similarity and disparity losses, only similarity loss, and only disparity), we train different feature extractors. We record the similarity of \ac{sp} and \ac{dp} feature descriptors during the training processes. The recording results in TABLE \ref{tab:ablation} indicate a decrease in the similarity of the \ac{sp} descriptors and an increase in the similarity of the \ac{dp} descriptors. The experimental findings highlight the necessity of the losses in the training of the illumination-robust feature extractor.

Except for the feature repeatability and similarity, we also test the accuracy of the homography estimation of the proposed illumination-robust feature extractor with the varying illumination images of HPatches \cite{balntas2017hpatches}. Setting the distance threshold $\epsilon = 3$, the estimation correctness of the proposed illumination-robust feature extractor (0.867) is comparable with SIFT (0.850) and SuperPoint (0.849), exceeding ORB (0.393) and D2-Net \cite{dusmanu2019d2} (0.558) significantly. These evaluation results indicate that the proposed illumination-robust feature extractor has the potential to used in common robotic applications, such as \ac{mvs} or other \ac{slam} applications. In this evaluation, the images containing objects or buildings perform higher estimation accuracy compared with images containing natural landscapes, which indicates that the application of relightable 3D reconstruction of outdoor landscapes would improve the performance of the illumination-robust feature extractor further. We illustrate some examples of the feature detection results in Fig. \ref{fig:matching_image}.


\section{Conclusion \& Future Work}\label{sec:conclusion}
This paper introduces a novel approach to address the challenge of feature repeatability and similarity under diverse illumination conditions. Our proposed method leverages the benefits of relightable 3D scene reconstruction-based data augmentation and self-supervised training frameworks. The relightable 3D scene reconstruction technique enables the establishment of a dataset containing RGB images in various illuminations. Subsequently, we formulate the training of illumination-robust feature extractor as a self-supervised framework by utilizing three losses, the repeatability loss, the similarity loss, and the disparity loss. Comprehensive experiments have been conducted to verify the effectiveness of the proposed method. The experiment results demonstrate that the feature extraction extractor trained with our data augmentation method can improve the feature detection repeatability and the similarity of \ac{sp} feature descriptors.

Future research should focus on continuously expanding the scalability and diversity of the dataset. More cross-domain feature training techniques can be applied to improve the performance of the illumination-robust feature extractor. Extensive tests should also be conducted to verify the performance of the illumination-robust feature extractor in real-world scenes, such as indoor and outdoor \ac{slam}.


\section*{Acknowledgment}
The authors would like to appreciate Jiawen Liu for her technical support in figure drawing.


\bibliographystyle{IEEEtran}
\bibliography{main}
\end{document}